\documentclass[10pt,twocolumn,letterpaper]{article}
\usepackage[pagenumbers]{cvpr} 
\makeatletter
\@namedef{ver@everyshi.sty}{}
\makeatother
\usepackage{epsfig}
\usepackage{graphicx}
\graphicspath{{img/}}
\usepackage{amssymb}
\usepackage[utf8]{inputenc} 
\usepackage[T1]{fontenc}    
\usepackage{url}            
\usepackage{amsfonts}       
\usepackage{nicefrac}       
\usepackage{microtype}      
\usepackage{xcolor}         
\usepackage{subcaption}
\usepackage[linesnumbered,ruled,vlined]{algorithm2e}
\usepackage{caption}
\usepackage{amsmath}
\usepackage{amsthm}
\usepackage{multirow}
\makeatletter
\def\blfootnote{\gdef\@thefnmark{}\@footnotetext}
\makeatother
\usepackage[pagebackref,breaklinks,colorlinks]{hyperref}
\hypersetup{colorlinks,linkcolor={blue},citecolor={blue},urlcolor={blue}}
\usepackage[capitalize]{cleveref}
\crefname{section}{Sec.}{Secs.}
\Crefname{section}{Section}{Sections}
\Crefname{table}{Table}{Tables}
\crefname{table}{Tab.}{Tabs.}
\usepackage{comment}
\usepackage{booktabs}
\usepackage{calc}
\usepackage{enumitem}
\begin{document}

\title{MATE: Masked Autoencoders are Online 3D Test-Time Learners}

\author{
M. Jehanzeb Mirza$^{\dagger1,2}$ \and 
Inkyu Shin$^{\dagger3}$ \and 
Wei Lin$^{\dagger1}$ \and
Andreas Schriebl$^1$ \and  
Kunyang Sun$^4$ \and  
Jaesung Choe$^3$ \and  
Mateusz Kozinski$^1$ \and 
Horst Possegger$^1$ \and 
In So Kweon$^3$ \and
Kuk-Jin Yoon$^3$ \and 
Horst Bischof$^{1,2}$\and\\
$^1$Institute for Computer Graphics and Vision, Graz University of Technology, Austria.\\
$^2$Christian Doppler Laboratory for Embedded Machine Learning.\\
$^3$Korea Advanced Institute of Science and Technology (KAIST), South Korea.\\
$^4$Southeast University, China.\\ 
}
\maketitle

\begin{abstract}
Our MATE is the first Test-Time-Training (TTT) method designed for 3D data, which makes deep networks trained for point cloud classification robust to distribution shifts occurring in test data.
Like existing TTT methods from the 2D image domain, 
MATE also leverages test data for adaptation. Its test-time objective is that of a Masked Autoencoder: a large portion of each test point cloud is removed before it is fed to the network, tasked with reconstructing the full point cloud. Once the network is updated, it is used to classify the point cloud. We test MATE on several 3D object classification datasets and show that it significantly improves robustness of deep networks to several types of corruptions commonly occurring in 3D point clouds. We show that MATE is very efficient in terms of the fraction of points it needs for the adaptation. It can effectively adapt given as few as 5\% of tokens of each test sample, making it extremely lightweight. Our experiments show that MATE also achieves competitive performance by adapting sparsely on the test data, which further reduces its computational overhead, making it ideal for real-time applications.
\end{abstract}
\blfootnote{
$\dagger$ Equally contributing authors.\\ 
{\indent\indent\phantom{.} Correspondence: \tt\small{muhammad.mirza@icg.tugraz.at}}
}
\begin{figure}
\vspace{0.5cm}
    \centering    \includegraphics{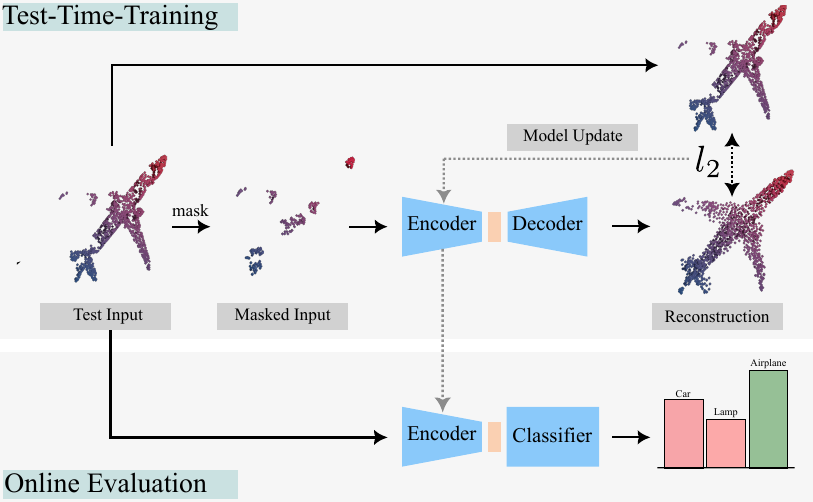}
    \caption{Overview of our Test-Time Training methodology. 
{
We adapt the encoder to a single out-of-distribution (OOD) test sample online by updating its weights using a self-supervised reconstruction task.
We then use the updated weights to make a prediction on the test sample.
}
{To enable this approach, the encoder, decoder, and the classifier are co-trained in the classification and reconstruction tasks~\cite{pang2022masked}, which is not shown in the figure.}
}
    \label{fig:teaser}
\end{figure}


\section{Introduction}
{
Recent deep neural networks show impressive performance in classifying 3D point clouds. 
However, their success is 
warranted only if the test data originates from the same distribution as training data.
In real-world scenarios, this assumption is often violated. 
A LiDAR point cloud can be corrupted, for example, due to sensor malfunction or environmental factors. 
It has been shown in~\cite{ren2022modelnet-c, sun2022benchmarking} that, even seemingly insignificant perturbations, like introduction of jitter or minute amount of noise to the point cloud, can significantly decrease the performance of several state-of-the-art 3D object recognition architectures.
This lack of robustness can limit the utility of 3D recognition in numerous applications, including in construction industry, geo-surveying, manufacturing and autonomous driving. 
Distribution shifts that can affect 3D data are diverse in nature and it might not be feasible to train the network for all the shifts which can possibly be observed in point clouds at test-time.
Thus, there is a need to adapt to these shifts online at test-time, in an unsupervised manner. 
}

{
Test-Time Training (TTT) leverages unlabeled test data to adapt the classifier to the change in data distributions at test-time in an online manner.
Several TTT approaches have been recently proposed for the 2D image domain.
The main techniques include regularizing the classifier on test data with objective functions defined on the entropy of its predictions~\cite{liang2020shot,wang2020tent,zhang2021memo},
updating the statistics of the batch normalization layers to match the distribution of the test data~\cite{mirza2022norm},
and training the network on test data with self-supervised tasks~\cite{liu2021tttpp,sun2020ttt}.
However, existing 2D TTT methods fail
when naively applied to the 3D point clouds, stressing upon the need for 3D-specific TTT methodologies, which are currently non-existent.
}

{
In this paper, we address the problem of test-time training for 3D point cloud classification.
We propose a 3D-specific method, MATE, which adopts the self-supervised paradigm~\cite{liu2021tttpp, sun2020ttt}, in which a deep network is adapted by solving a self-supervised task for the OOD test data.
Our choice is dictated by the availability of a self-supervised task that perfectly matches our goal of adapting 3D networks. 
Masked autoencoder proved very effective in pre-training 3D object recognition networks~\cite{pang2022masked}, and adapting deep networks to corruptions of 2D images~\cite{gandelsman2022test}.
It removes a large portion of the point cloud, and tasks the network with reconstructing the entire point cloud given only the part that has not been removed.
We use this procedure to update the network on every test sample that is used for the adaptation.
An overview 
is provided in Figure~\ref{fig:teaser}.
}

Our main contributions 
are extending TTT to the 3D point cloud domain and showing that simply adopting TTT techniques widely used in the 2D image domain is not a viable solution for 3D, stressing out the need for 3D-specific approaches.
To this end, we demonstrate how well-suited and powerful masked autoencoding is to address online test-time training for 3D data.
We conduct extensive evaluations on three point cloud recognition datasets.
Apart from achieving strong performance gains for online adaptation, we discover and highlight several useful properties for TTT with masked autoencoders.
For example, our MATE achieves significant performance gains even when masking 95\% of tokens from the point clouds. 
This seemingly nuance can have important benefits: At test-time, the encoder only needs to process the remaining 5\% of the visible tokens to adapt the network, radically limiting the computational overhead of the adaptation. 
The overhead from TTT can be further reduced by adapting sparsely to test data, as MATE can achieve significant performance gains over un-adapted networks by only adapting on every 100-th sample of the OOD test data.

\section{Related works}
Our work is related to Unsupervised Domain Adaptation (UDA), Self-Supervised Learning (SSL) and more closely to methods which learn on test instances. 

\paragraph{Unsupervised Domain Adaptation.} UDA methods aim to bridge the domain gap between the source and target domains without requiring access to labels from the target domain. 
UDA has gained considerable traction in the 3D vision community. 
PointDAN~\cite{qin2019pointdan} aligns local and global point cloud features from the source and target domain in an end-to-end manner. 
Liang~\etal~\cite{liang2022point} propose to predict masked local structures by estimating cardinality, position and normals for the point cloud.
Shen~\etal~\cite{shen2022domain} first propose to encode the underlying geometry of point clouds from the target data with the help of implicit functions and resort to pseudo-labeling in the second step.
For 3D object detection, adversarial augmentation is proposed by 3D-VField~\cite{shen2022domain} for generalization to different domains.
MLC-Net~\cite{luo2021unsupervised} proposes to use a student-teacher network along with pseudo-labeling. 
Wang~\etal~\cite{wang2020train} propose to bridge the domain gap for 3D object detection by using priors, such as bounding box sizes from the target domain. 
Although unsupervised domain adaptation approaches tackle an important problem, they assume knowledge about the test distribution and try to mitigate the distribution mismatch by an extensive training phase. 
On the other hand, test-time training requires no such priors and offers a setting which is more closer to real world scenarios, where on-the-fly adaptation is required. 
\paragraph{Self-Supervised Learning.}
Self-Supervised representation learning thrives on the idea of extracting supervision from the data itself. 
A popular SSL training objective is to bring the representations from the two randomly augmented views from the same sample closer and push apart the views from the other samples in the batch~\cite{chen2020simple, chen2021exploring, huang2021spatio, zbontar2021barlow}. 
Another approach for SSL is to extract the supervision from the reconstruction of the input data.
Self-supervised representation learning by using Autoencoders~\cite{vincent2008extracting} has been a long-standing research topic in computer vision.
Recently, He~\etal~\cite{he2022masked} 
proposed Masked Autoencoders~(MAE) for self-supervised representation learning in the image domain.
MAE uses an asymmetric encoder-decoder structure based on the Vision Transformer~\cite{dosovitskiy2020image}.
High proportion of the image tokens~($70-75\%$) are masked and the SSL objective is to reconstruct the masked tokens.
On a similar note, Pang~\etal~\cite{pang2022masked} propose Point-MAE, an MAE framework for self-supervised representation learning in 3D point cloud domain and show that due to the sparse nature of point clouds, a more severe masking ratio can also be employed. 
In our work we also use reconstruction of point clouds as an auxiliary self-supervised task for test-time training. 
To this end, we use the PointMAE framework and at test-time get our supervisory signal by reconstructing highly masked regions from the OOD input point cloud. 
\begin{figure*}
    \centering
    \includegraphics[width=1\textwidth]{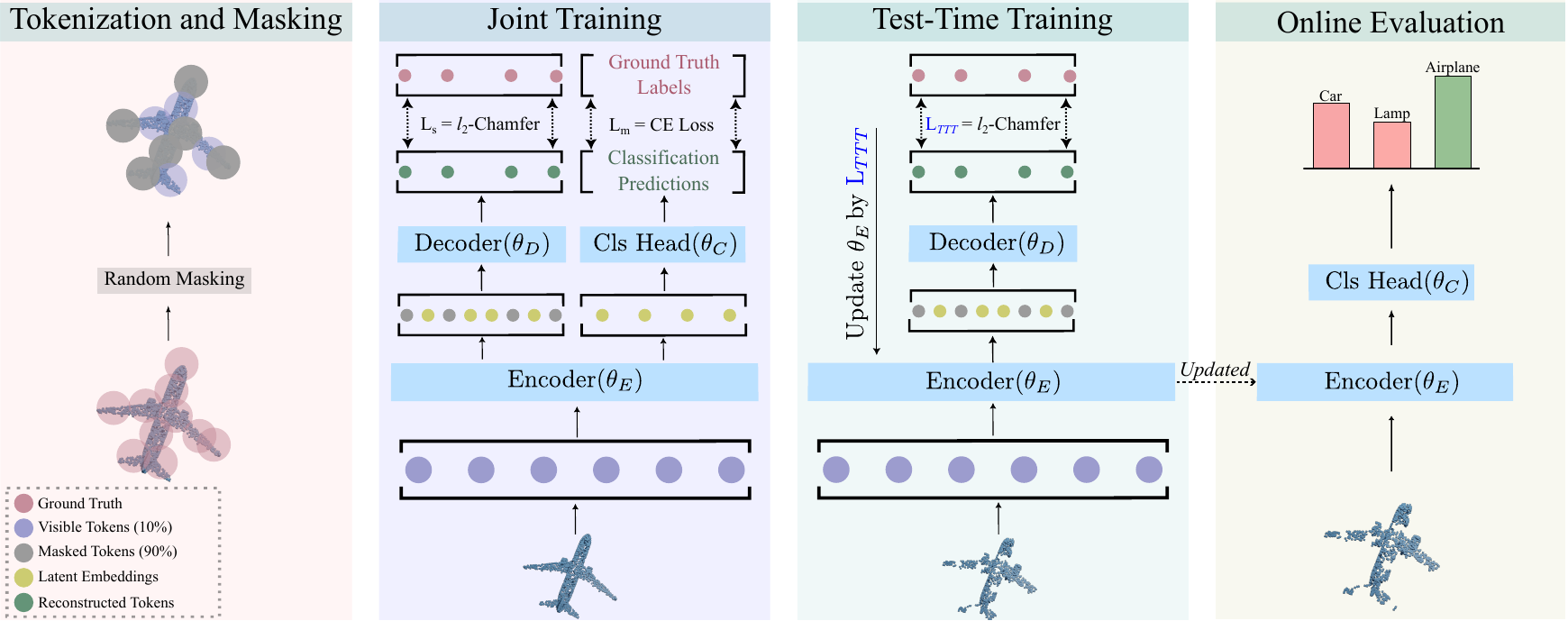}
    \caption{Overview of our 3D Test-Time Training methodology. 
    We build on top of PointMAE. The input point cloud is first tokenized and then randomly masked. For our setup, we mask $90\%$ of the point cloud. For joint training the visible tokens from the training data are fed to the encoder to get the latent embeddings from the visible tokens. These embeddings are fed to the classification head for the classification loss and concatenated with the masked tokens and fed to the decoder for reconstruction to obtain the reconstruction loss. Both losses are optimized jointly. For adaptation to an out-of-distribution test sample at test-time, we only use the MAE reconstruction task. Finally, after adapting the encoder on this single sample, evaluation is performed by using the updated encoder weights.
        }
    \label{fig:method}
\end{figure*}

\paragraph{Test-Time Training.}
TTT methods can be divided in to two distinct groups. 
The first group of methods
add post-hoc regularization for adaptation to OOD test data.
Boudiaf~\etal~\cite{boudiaf2022parameter} propose a gradient free TTT approach, which promotes consistency of output predictions coupled with Laplacian regularization.
TENT~\cite{wang2020tent}, SHOT~\cite{liang2020shot}~and~MEMO~\cite{zhang2021memo} rely on entropy minimization from the output softmax distribution.
T3A~\cite{iwasawa2021t3a} casts TTT as a prototype learning problem, while DUA~\cite{mirza2022norm} employs online statistical correction in the batch normalization layers for TTT.
We test several of these approaches by porting them for TTT in the 3D point cloud recognition task but none of these approaches prove to be a competitive baseline for our MATE (Section.~\ref{subsec:results}), further highlighting the need for 3D-specific methods. 

The other group of methods propose to use auxiliary self-supervised tasks for adaptation to distribution shifts at test-time and are more closely linked to our MATE. 
Sun~\etal~\cite{sun2020ttt} employ rotation prediction~\cite{gidaris2018unsupervised} as an auxiliary task for TTT. 
TTT++~\cite{liu2021tttpp} uses contrastive self-supervised learning (SimCLR~\cite{chen2020simple}) as an auxiliary objective.
TTT-MAE~\cite{gandelsman2022test} substitutes the self-supervised objective with Masked Autoencoder~\cite{he2022masked} reconstruction task for TTT in the image domain.
A general insight from these works implies that the choice of auxiliary self-supervised task is of utmost importance.
{MATE also employs the task of masked auto-encoding to drive the adaptation, but it reconstructs point clouds instead of images.
This forces the network to encode the geometry of the point cloud and model long-range dependencies between local shapes.
Furthermore, our experiments show that, for 3D point clouds, geometric reconstruction is a better auxiliary task than rotation prediction, which is employed by TTT~\cite{sun2020ttt}.
}

\section{MATE}
We first describe our problem setting and model architecture in detail, then we describe our training setup and finally provide details about our test-time training methodology.
\subsection{Problem setting}
We follow the conventional test-time training setting, proposed by TTT~\cite{sun2020ttt}, where at test-time we first adapt on a single sample and then test it.  
For adaptation we use the MAE reconstruction task.
To process the point clouds, we use the PointMAE~\cite{pang2022masked}. 
{Given a point cloud~$\mathcal{X}{=}\{\mathbf{p}_i\}^N_{i=1}$ of $N$ points $\mathbf{p}_i {=} (x, y, z)^T$,
the points are grouped into tokens, that is, possibly overlapping subsets of nearby points, using the farthest point sampling~\cite{pang2022masked}.
A proportion of tokens equal to the mask ratio~$m$ is then randomly masked, yielding the   
masked tokens, 
that we denote by $\mathcal{X}^m$, while $\mathcal{X}^v$ represent the remaining visible tokens.}
During {joint} training, we assume access to the training data $\mathcal{S}=\{(\mathcal{X}, \mathcal{Y})\}$, where each point cloud $\mathcal{X}$ is accompanied by its ground truth label~$\mathcal{Y}$.
During test-time training, we do not have access to the entire test dataset but instead adapt to each single sample as it is encountered.
After adapting the network parameters on each sample, the updated weights are used for predicting the class label. 
A detailed overview of different stages in our pipeline is shown in Figure~\ref{fig:method}, while the pseudocode is provided in the supplementary material.
\subsection{Architecture}
{
We adopt the PointMAE architecture~\cite{pang2022masked}, proven to work well in unsupervised pre-training for 3D object classification.
It consists of an encoder $E$, a decoder $D$, a prediction head $P$, and a classifier head $C$.}
The encoder $E$ consists of $12$ standard transformer blocks and receives only the unmasked point patches as input. 
The decoder $D$ is similar to $E$, however, it is lightweight ($4$ blocks), which makes the encoder-decoder structure asymmetrical. 
The masked point patches and the embeddings from the unmasked point patches are fed to the decoder after concatenation. 
The decoder feeds the embeddings to the prediction head $P$, which is a simple linear fully connected layer and reconstructs the points in coordinate space.
The classifier head $C$ is a projection 
from the dimensions of the encoder output to the number of classes in the respective dataset. 
We use $3$ fully connected layers with ReLU non-linearity, batch normalization and dropout as our classification head. 

\subsection{Joint Training}
\label{sec:jointTraining}
{
Previous methods that employ the masked autoencoder for images or point clouds~\cite{gandelsman2022test, pang2022masked} pre-train the encoder and decoder in a self-supervised manner and subsequently train the classifier on top of it.
In contrast, to make the encoder learn embeddings that at the same time describe the input geometry and are well suited for the downstream task, we train the two heads jointly.
}
Given all the parameters of the network~$\{\theta_E,\,\theta_D,\,\theta_P,\,\theta_C\}$, the joint training is posed as
\begin{multline}
\min_{\theta_E,\,\theta_D,\,\theta_P,\,\theta_C} 
    \mathbb{E}_{(\mathcal{X},\mathcal{Y})\in\mathcal{S}} \big[ L_{c}(\mathcal{X},\mathcal{Y};\theta_E, \theta_C)   \\
        + \lambda \cdot L_{s} (\mathcal{X};\theta_E, \theta_D, \theta_P ) \big],
\end{multline}
where the expectation is taken over the training set $\mathcal{S}$, and the hyper-parameter $\lambda$ balances the two tasks. 
We set $\lambda=1$ for all experiments.
Here, $L_{c}$ is a cross entropy (CE) loss to learn the main classification task 
\begin{equation}
L_{c}(\mathcal{X},\mathcal{Y};\theta_E, \theta_C)=CE(C \circ E(\mathcal{X}^v), \mathcal{Y}),
\end{equation}
where $\mathcal{X}^v$ are the visible tokens and $L_{s}$ is the self-supervised loss.
Following~\cite{pang2022masked}, we use 
\begin{equation}
\label{eq:chamfer-loss}
L_{s}(\mathcal{X};\theta_E, \theta_D, \theta_P )=CD(P \circ D \circ E(\mathcal{X}^{v}), \mathcal{X}),
\end{equation}
which is the Chamfer distance $CD$ between the reconstructed tokens, and the training point sets~$\mathcal{X}$.


\subsection{Test-Time Training}
Given the parameters $\{\theta_E,\,\theta_D,\,\theta_P,\,\theta_C\}$, trained jointly for the main classification task and the self-supervised reconstruction task on the training data. Our goal at test-time is to adapt to the OOD
test data in an unsupervised manner, to achieve generalization.
For this purpose we use the self-supervised MAE reconstruction task to adapt the network parameters to the OOD test sample.  

For adaptation at test-time, we are granted access to only a single out-of-distribution point-cloud $\tilde{\mathcal{X}}$, without any ground truth label.
The point cloud is tokenized and masked, and processed by the encoder~$E$ which yields the encoding vector. 
Finally, the patch encodings and the masked patches are concatenated and fed to the decoder~$D$ and ultimately to the prediction head~$P$ to obtain the reconstructed point cloud.
The reconstruction loss is again an $l_2$ Chamfer distance between the reconstructed masked tokens and the corresponding ground truth tokens from the original out-of-distribution test sample. 
Our objective at test-time is to update the parameters of the encoder $\theta_E$, decoder $\theta_D$ and the prediction head $\theta_P$ to generalize to the OOD test sample.
More formally, for test-time training we minimize 
\begin{equation}
\label{eq:loss-ttt}
L_{TTT} = \min_{\theta_E, \theta_D, \theta_P} L_{s}(\tilde{\mathcal{X}} ;\theta_E, \theta_D, \theta_P).
\end{equation}
Although for the downstream task of object classification, we only require the updated encoder, through experiments we find that updating the decoder and the prediction head does not affect the final classification performance. 

\subsection{Online Adaptation Variants}
After adapting the encoder weights by the reconstruction loss during test-time training, prediction scores for the OOD sample are obtained by using the classifier head~$C$, from the joint training phase.  
Following TTT~\cite{sun2020ttt}, we provide two variants of our MATE, which are described as follows:
\paragraph{MATE-Standard} only assumes access to a single point cloud sample at test-time and the goal is to iteratively adjust the weights on single samples in order to make the right prediction. 
For this purpose, we 
perform 20 gradient steps on the encoder parameters $\theta_E$ to minimize the objective in Eq.~\eqref{eq:loss-ttt}, computed for one test sample.
As the next sample is received, we reinitialize the weights for all the parameters $\{\theta_E, \theta_D, \theta_P\}$, and repeat the same process again.
\paragraph{MATE-Online} assumes that point clouds are received in a stream. 
For this version, we accumulate the model updates after adaptation on each sample. 
We only calculate (and backpropagate) $L_{TTT}$ from Eq.~\eqref{eq:loss-ttt}, once for each sample. 


\subsection{Augmentations}
During joint-training we only train the network with point cloud scale and translation augmentations, as originally used by 
 the authors of PointMAE.
For test-time training, we do not use any augmentation, instead we construct a batch~(following~\cite{sun2020ttt}) from the single point cloud sample and for reconstruction, we randomly mask~$90\%$ of the tokens. 
 Random masking is essential for MAE and also provides us with a natural augmentation.
 We further find that we can increase the masking ratio up to~$95\%$ and still get an impressive performance improvement. 
 This is in contrast to images where a masking ratio of up to~$70-75\%$ is employed. 
 Higher masking ratios help in efficient test-time training,
 since only the unmasked  tokens are processed by the encoder, which carries the majority of the computation effort because it has a larger structure than the decoder. 
 
\section{Experimental Evaluation}
We provide results for both the Standard and the Online evaluation variants.
Here, we first describe the datasets we use for evaluation, second we provide our implementation details and later present our results. 
\subsection{Datasets}
We test MATE on the task of object classification for 3D point clouds. 
To this end, we use $3$ popular object classification datasets.


\paragraph{ModelNet-40C.}
ModelNet-40C~\cite{sun2022benchmarking} is a benchmark for evaluating robustness of point cloud classification architectures.
In this benchmark, $15$ common types of corruptions are {induced} on the original test set of ModelNet-40~\cite{wu20153d}. 
These corruptions are divided into $3$ parent categories comprising \emph{transformation}, \emph{noise} and \emph{density}.
Their goal is to mimic distribution shifts which occur in real-world,~\eg, common noise patterns on a LiDAR scan due to fault in the sensors capturing the data.

\paragraph{ShapeNet-C.}
ShapeNetCore-v2~\cite{chang2015shapenet} is a large-scale point cloud classification dataset consisting of $51127$ shapes from $55$ categories. 
We divide this dataset into three splits, train ($35789$, $70\%$), validation (5113, $10\%$) and test ($10225$, $20$\%).
We provoke $15$ different corruptions in the test set of ShapeNet, similar to ModelNet-40C, by using the open source implementation provided by~\cite{sun2022benchmarking}.
We refer to this dataset as ShapeNet-C.

\paragraph{ScanObjectNN-C.}
ScanObjectNN~\cite{uy-scanobjectnn-iccv19} is a point cloud classification dataset which is collected in the real-world. 
It consists of $15$ categories with $2309$ samples in the train set and $581$ samples in the test set. 
We again use the open source code provided by~\cite{sun2022benchmarking} to cause $15$ different corruptions in the test set of ScanObjectNN for our evaluations, which we refer to as ScanObjectNN-C. 

\begin{table*}
\setlength\tabcolsep{2.0pt}
\centering
    \small
\begin{tabular}{lccccccccccccccc|cc}
      \multicolumn{1}{r}{ \rotatebox{0}{corruptions:}}            &  \rotatebox{0}{uni} &  \rotatebox{0}{gauss}  &  \rotatebox{0}{backg} &  \rotatebox{0}{impul} &  \rotatebox{0}{upsam}  &   \rotatebox{0}{rbf}  &   \rotatebox{0}{rbf-inv}   &  \rotatebox{0}{den-dec} & \rotatebox{0}{dens-inc}  &   \rotatebox{0}{shear}   &  \rotatebox{0}{rot}   &   \rotatebox{0}{cut}  &   \rotatebox{0}{distort}  &   \rotatebox{0}{oclsion}   &    \rotatebox{0}{lidar}  &   \rotatebox{0}{Mean}  & \\

\midrule
Source-Only &             {66.6} &             {59.2} &              {7.2} &             {31.7} &             {74.6} &             {67.7} &             {69.7} &             {59.3} &             {75.1} &             {74.4} &             {38.1} &             {53.7} &             {70.0} &             {38.6} &             {23.4} &             {53.9} \\
Joint-Training &             {62.4} &             {57.0} &             {32.0} &             {58.8} &             {72.1} &             {61.4} &             {64.2} &             {75.1} &             {80.8} &             {67.6} &             {31.3} &             {70.4} &             {64.8} &             {36.2} &             {29.1} &             {57.6} \\
           DUA &             {65.0} &             {58.5} &             {14.7} &             {48.5} &             {68.8} &             {62.8} &             {63.2} &             {62.1} &             {66.2} &             {68.8} & {\underline{46.2}} &             {53.8} &             {64.7} & {\underline{41.2}} & {\underline{36.5}} &             {54.7} \\
           TTT-Rot &             {61.3} &             {58.3} &  {\bfseries{34.5}} &             {48.9} &             {66.7} &             {63.6} &             {63.9} &             {59.8} &             {68.6} &             {55.2} &             {27.3} &             {54.6} &             {64.0} &             {40.0} &             {29.1} &             {53.0} \\
          SHOT &             {29.6} &             {28.2} &              {9.8} &             {25.4} &             {32.7} &             {30.3} &             {30.1} &             {30.9} &             {31.2} &             {32.1} &             {22.8} &             {27.3} &             {29.4} &             {20.8} &             {18.6} &             {26.6} \\
           T3A &             {64.1} &             {62.3} & {\underline{33.4}} &             {65.0} &             {75.4} &             {63.2} &             {66.7} &             {57.4} &             {63.0} &             {72.7} &             {32.8} &             {54.4} &             {67.7} &             {39.1} &             {18.3} &             {55.7} \\
          TENT &             {29.2} &             {28.7} &             {10.1} &             {25.1} &             {33.1} &             {30.3} &             {29.1} &             {30.4} &             {31.5} &             {31.8} &             {22.7} &             {27.0} &             {28.6} &             {20.7} &             {19.0} &             {26.5} \\
        \midrule
 MATE-Standard & {\underline{75.0}} & {\underline{71.1}} &             {27.5} & {\underline{67.5}} & {\underline{78.7}} & {\underline{69.5}} & {\underline{72.0}} &  {\bfseries{79.1}} & {\underline{84.5}} & {\underline{75.4}} &             {44.4} & {\underline{73.6}} & {\underline{72.9}} &             {39.7} &             {34.2} & {\underline{64.3}} \\
   MATE-Online &  {\bfseries{82.9}} &  {\bfseries{80.6}} &             {32.4} &  {\bfseries{74.0}} &  {\bfseries{85.7}} &  {\bfseries{78.3}} &  {\bfseries{80.2}} & {\underline{78.1}} &  {\bfseries{86.5}} &  {\bfseries{79.3}} &  {\bfseries{56.6}} &  {\bfseries{77.9}} &  {\bfseries{77.1}} &  {\bfseries{49.7}} &  {\bfseries{50.0}} &  {\bfseries{71.3}} \\

\midrule
\end{tabular}
\caption{Top-1 Classification Accuracy (\%) for all distribution shifts in the ModelNet-40C dataset. All results are  for the PointMAE backbone trained on clean train set and adapted to the OOD test set with a batch-size of $1$. \emph{Source-Only} denotes its performance on the corrupted test data without any adaptation. Highest Accuracy is in bold, while second best is underlined.}
\label{tab:modelnet-c-results}
\end{table*}

\subsection{Implementation Details}
We jointly train a network for supervised classification and self-supervised reconstruction tasks, {as described in Section~\ref{sec:jointTraining}}. 
For joint training we only use $10\%$ of the visible tokens for the self-supervised reconstruction and the classification task. 
However, to obtain the final classification scores {at test-time,} we always feed $100\%$ of the tokens to the PointMAE backbone. 
For ModelNet-40 and ShapeNetCore experiments, we train the networks from scratch for $300$ epochs with a learning rate of $0.001$ and Cosine scheduler.
ScanObjectNN is a small-scale dataset, thus, we finetune the PointMAE network pre-trained on the large-scale ShapeNet-55~\cite{chang2015shapenet} dataset with a learning rate of $0.0005$ and a Cosine scheduler for only $100$ epochs, to avoid overfitting. 
All these models (including the vanilla PointMAE) use only the point cloud scaling and translation as augmentations\footnote{We avoid other augmentations,~\eg~jitter or rotation, because they might correlate with the corruptions in the ModelNet-C benchmark and can provide us with an unfair advantage during TTT.}.
For a fair comparison, the architectural details for all baselines and our method are kept constant.  

During test-time training we update the encoder, decoder and the prediction head only.
The classification head remains frozen. 
We use a learning rate of $5\mathrm{e}{-5}$ for TTT on ModelNet-40C, a learning rate of $1\mathrm{e}{-4}$ for ShapeNet-C and ScanObjectNN-C.
We use AdamW optimizer for both, pre-training and the test-time training. 
To calculate the test-time training loss, we construct a batch of $48$ from the single corrupted point cloud at test-time and randomly mask $90\%$ of each sample in the batch.
To encourage reproducibility, our entire codebase and pre-trained models are available at this repository: \href{https://github.com/jmiemirza/MATE}{https://github.com/jmiemirza/MATE}.

\subsection{Baselines}
\noindent
We compare our MATE to several other TTT approaches proposed for images. 
In our work we assume access to only a single sample for adaptation at test-time, thus, for a fair comparison with our MATE, we also test other baselines in the single sample adaptation protocol.
However, many 2D baselines fail in the single sample protocol, thus, we also provide results for larger batch sizes. 
A brief description of all the baselines is as follows.
\begin{itemize}[nosep,label=-,wide]
\item \emph{Source Only} refers to the PointMAE backbone trained in a supervised manner on the classification task only.
For testing on the OOD data, we do not mask the tokens, instead feed the entire point cloud. 
\item \emph{Joint Training}~\cite{hendrycks2019using} results are obtained by training the network jointly on the classification and MAE reconstruction task and testing it on the target data (\eg ModelNet-40C) without adaptation. 
\item \emph{SHOT}~\cite{liang2020shot} proposes to minimize the expected entropy of predictions calculated from the output probability distribution from the network. 
\item \emph{T3A}~\cite{iwasawa2021t3a} relies on learning class specific prototypes to replace the classifier which is learned on the training set. 
\item \emph{TENT}~\cite{wang2020tent} also minimizes the entropy of predictions from the output of the classifier. 
\item \emph{DUA}~\cite{mirza2022norm} updates the batch normalization statistics to adapt to OOD test images at test-time.
\item \emph{TTT-Rot}~\cite{sun2020ttt} with self-supervised rotation prediction task proposes to adapt to test data at test-time by predicting the rotation of images. 
Following the original paper, we train a network for classification and rotation prediction tasks. 
\end{itemize}

\begin{table}[]
    \setlength{\tabcolsep}{4pt}
    \centering
    \begin{tabular}{c|ccccc}
    Method&Source&TENT&SHOT&T3A&MATE-O\\
         \hline
         Accuracy (\%)&\multirow{ 2}{*}{53.9}&\multirow{ 2}{*}{65.6}&\multirow{ 2}{*}{63.8}&\multirow{ 2}{*}{55.9}&\multirow{ 2}{*}{74.5}\\
         (BS - 128)&&&&\\
        \bottomrule
    \end{tabular}
    \caption{Mean Top-1 Classification Accuracy (\%) for ModelNet-40C by using a larger batch size (BS) of $128$ for baselines and MATE-Online.}
    \label{tab:larger-bs-results}
\end{table} 
\subsection{Results}
\label{subsec:results}
\paragraph{ModelNet-40C:} In Table~\ref{tab:modelnet-c-results} we provide the results for all the distribution shifts in the ModelNet-40C dataset.
From the table, we see that our MATE outperforms other baselines comfortably.
Furthermore, even our MATE-Standard performs better than the baselines with a considerable margin, while also performing favorably on individual distribution shifts.
The test-time training approaches which rely on post-hoc regularization,
~\eg SHOT~\cite{liang2020shot} and TENT~\cite{wang2020tent} perform poorly, while T3A~\cite{iwasawa2021t3a} is only marginally above Source-Only baseline. 
This shows that the approaches designed for image data cannot be trivially transferred to the 3D domain. 
Moreover, all these approaches require larger batch sizes to work in the 2D domain. 
These approaches cannot adapt on a single test sample at test-time. 
For example, the entropy based approaches~\cite{liang2020shot, wang2020tent}, can have a trivial solution while optimizing the entropy of a single 
test sample. 
For larger batch sizes, we see that SHOT, TENT and T3A show some improvement in results (Table~\ref{tab:larger-bs-results}) but still MATE outperforms them comfortably.  
However, we reason that in online real-time applications we cannot access a batch of test data for adaptation, thus it is necessary that the TTT approaches work well even while having access to a single sample for adaptation at test-time.

From the results we also see that the mean performance over all corruptions of TTT-Rot falls below Source-Only, even though it is originally designed for the single sample adaptation scenario in the 2D domain. 
This could be an indication that the rotation prediction task is not well suited for test-time adaptation for 3D data. 
However, for Background corruption TTT-Rot~\cite{sun2020ttt} fares well. 
This might be because Background corruption introduces artifacts in the background
and TTT-Rot uses the entire point cloud for test-time adaptation, so it can adapt to this corruption better. 
On the other hand, we only adapt with $10\%$ of the visible tokens and might not be able to capture these artifacts introduced in the background.
Furthermore, we analyze the reconstructions from the background corruption and find that the reconstruction results are worse as compared to other corruptions. 
We show these visualizations in the supplemental. 
These reconstruction results suggest that the reconstruction task is co-related with the classification task.
Hence, better reconstruction accounts for better adaptation performance. 
We also see a similar trend for the TTT loss and classification accuracy at each adaptation step for corruptions in the ModelNet-40C.
These results are also delegated to the supplementary material. 
\begin{table*}
\setlength\tabcolsep{2.0pt}
\centering
    \small
\begin{tabular}{lccccccccccccccc|cc}
      \multicolumn{1}{r}{ \rotatebox{0}{corruptions:}}            &  \rotatebox{0}{uni} &  \rotatebox{0}{gauss}  &  \rotatebox{0}{backg} &  \rotatebox{0}{impul} &  \rotatebox{0}{upsam}  &   \rotatebox{0}{rbf}  &   \rotatebox{0}{rbf-inv}   &  \rotatebox{0}{den-dec} & \rotatebox{0}{dens-inc}  &   \rotatebox{0}{shear}   &  \rotatebox{0}{rot}   &   \rotatebox{0}{cut}  &   \rotatebox{0}{distort}  &   \rotatebox{0}{oclsion}   &    \rotatebox{0}{lidar}  &   \rotatebox{0}{Mean}  & \\

\midrule
    Source-Only &             {69.2} &             {62.8} &             {10.3} &             {56.2} &             {70.1} &             {70.5} &             {71.9} & {\underline{85.5}} & {\underline{86.2}} &             {73.9} &             {41.3} & {\underline{84.4}} &             {69.9} &              {7.9} &              {3.9} &             {57.6} \\
Joint-Training &             {72.5} &             {66.4} &             {15.0} &             {60.6} &             {72.8} &             {72.6} &             {73.4} &             {85.2} &             {85.8} &             {74.1} &             {42.8} &             {84.3} &             {71.7} &              {8.4} &              {4.3} &             {59.3} \\
           DUA &             {76.1} &             {70.1} &             {14.3} &             {60.9} &             {76.2} &             {71.6} &             {72.9} &             {80.0} &             {83.8} &             {77.1} & {\underline{57.5}} &             {75.0} &             {72.1} &             {11.9} &             {12.1} &             {60.8} \\
       TTT-Rot &             {74.6} &             {72.4} & {\underline{23.1}} &             {59.9} &             {74.9} &             {73.8} &             {75.0} &             {81.4} &             {82.0} &             {69.2} &             {49.1} &             {79.9} &             {72.7} & {\underline{14.0}} &             {12.0} &             {60.9} \\
          SHOT &             {44.8} &             {42.5} &             {12.1} &             {37.6} &             {45.0} &             {43.7} &             {44.2} &             {48.4} &             {49.4} &             {45.0} &             {32.6} &             {46.3} &             {39.1} &              {6.2} &              {5.9} &             {36.2} \\
           T3A &             {70.0} &             {60.5} &              {6.5} &             {40.7} &             {67.8} &             {67.2} &             {68.5} &             {79.5} &             {79.9} &             {72.7} &             {42.9} &             {79.1} &             {66.8} &              {7.7} &              {5.6} &             {54.4} \\
          TENT &             {44.5} &             {42.9} &             {12.4} &             {38.0} &             {44.6} &             {43.3} &             {44.3} &             {48.7} &             {49.4} &             {45.7} &             {34.8} &             {48.6} &             {43.0} &             {10.0} &             {10.9} &             {37.4} \\
        \midrule
 MATE-Standard & {\underline{77.8}} & {\underline{74.7}} &              {4.3} & {\underline{66.2}} & {\underline{78.6}} & {\underline{76.3}} & {\underline{75.3}} &  {\bfseries{86.1}} &  {\bfseries{86.6}} & {\underline{79.2}} &             {56.1} &             {84.1} & {\underline{76.1}} &             {12.3} & {\underline{13.1}} & {\underline{63.1}} \\
   MATE-Online &  {\bfseries{81.5}} &  {\bfseries{78.6}} &  {\bfseries{40.9}} &  {\bfseries{75.9}} &  {\bfseries{81.6}} &  {\bfseries{79.7}} &  {\bfseries{80.1}} &             {84.9} &             {85.9} &  {\bfseries{81.8}} &  {\bfseries{70.8}} &  {\bfseries{85.1}} &  {\bfseries{79.0}} &  {\bfseries{14.2}} &  {\bfseries{16.6}} &  {\bfseries{69.1}} \\

\midrule
\end{tabular}
\caption{Top-1 Classification Accuracy (\%) for all distribution shifts in the ShapeNet-C dataset. All results are  for the PointMAE backbone trained on clean train set set and adapted to the OOD test set with a batch-size of 1. 
}
\label{tab:shapenet-c-results}
\end{table*}

\begin{table}[]
    \centering
    \small
    \begin{tabular}{lc c lc}
    \multirow{1}{*}{Method} & Accuracy (\%) && \multirow{1}{*}{Method} &  Accuracy (\%)\\
    \cmidrule{1-2}
    \cmidrule{4-5}
         Source&45.7 && TTT-Rot&46.1\\
         SHOT&38.3 &&T3A&40.3\\
       JT  &45.6 &&MATE-S&\underline{47.0}\\
       
       DUA&46.0&&MATE-O&\textbf{48.5}\\
       \cmidrule{1-2}
       \cmidrule{4-5}
    \end{tabular}
\caption{Top-1 Classification Accuracy (\%) averaged over the 15 corruptions in the ScanObjectNN-C dataset (adapted with batch size 1). JT: Joint Training, MATE-S: MATE-Standard, MATE-O:~MATE-Online   }
    \label{tab:scanobjectnn-results}
\end{table}
\begin{table}[]

    \centering
    \small
    \begin{tabular}{lcccccc}
         &\multicolumn{6}{c}{Mask Ratio (\%)}  \\
         \midrule
         & 97.5&95&90&80&70&60\\
         \midrule
         MATE&\multirow{ 2}{*}{56.9}&\multirow{ 2}{*}{71.6}&\multirow{ 2}{*}{71.3}&\multirow{ 2}{*}{71.5}&\multirow{ 2}{*}{71.6}&\multirow{ 2}{*}{71.5}\\
         Online\\
         \midrule
    \end{tabular}
    \caption{Top-1 Classification Accuracy (\%) averaged over all corruptions in the ModelNet-40C dataset, while using different masking ratios for test-time training. 
    The accuracy for Source-Only baseline is 57.6\%. }
    \label{tab:ablation-masking-ratio}
    \vspace{-0.3cm}
\end{table}

\paragraph{ShapeNet-C:} In Table~\ref{tab:shapenet-c-results} we provide Top-1 Accuracy (\%) for object classification on the ShapeNet-C dataset. 
We again see that both evaluation variants of our MATE show impressive results on the large-scale ShapeNet dataset.
MATE-Online has a huge performance gain over other baselines, which is expected, since for these evaluations we accumulate the model updates.
Similarly, MATE-Standard also outperforms other baselines and even surpasses MATE-Online on the density-related corruptions of the point clouds. 
We again notice that popular 2D test-time training methods~\cite{iwasawa2021t3a,liang2020shot, mirza2022norm,sun2020ttt, wang2020tent} struggle for the ShapeNet dataset as well. 
These results further strengthen our reasoning that the need for 3D test-time training cannot be fulfilled by naively porting the 2D TTT approaches.  

\paragraph{ScanObjectNN-C:} We also test our MATE on point clouds collected in real world, on which we introduce the corruptions proposed in the ModelNet-C benchmark~\cite{sun2022benchmarking}. 
The results are provided in Table~\ref{tab:scanobjectnn-results} and are in-line with the other datasets.
These results show the applicability of MATE on data collected in the real world scenarios as well. 

\section{Ablation Studies}
We additionally test how MATE performs with different masking ratios, scenarios where sparse adaptation on test samples is required, the effect of batch size on TTT and the effect on performance while combining multiple corruption types together. 

\subsection{Masking Ratios}
\label{subsec:masking-ratio}
The PointMAE has an asymmetric encoder-decoder design. 
The decoder is a lightweight architecture, while the encoder is a deeper network.
Therefore, most of the computation effort is spent in the encoding part of the pipeline.
Since the encoder processes only the visible tokens, higher masking ratio implies lower burden for the encoder.
We find that our MATE can work with extremely high masking ratios, making test-time training very efficient. 
The results for adaptation with different masking ratios are provided in Table~\ref{tab:ablation-masking-ratio}.
We see that even with a severe masking of~$95\%$ of the tokens (\ie only processing~5\% visible tokens), our MATE can achieve~$14$ percent-points over the Source-Only (without adaptation) results.
Even with $97.5\%$ masking, we still improve on the Source-Only results. 
These results also show that lower masking ratios do not give us more gain in performance 
but instead could induce latency during test-time training, undesirable for real-time applications.

\begin{table}[]
\setlength\tabcolsep{3.8pt}

    \centering
    \small
    \begin{tabular}{lcccccccc}
         &\multicolumn{8}{c}{Batch Size for Test-Time Training}  \\
         \midrule
         & 1&2&8&16&24&32&40&48\\
         \midrule
         MATE&\multirow{ 2}{*}{43.1}&\multirow{2}{*}{66.4}&\multirow{ 2}{*}{69.7}&\multirow{ 2}{*}{70.2}&\multirow{ 2}{*}{70.4}&\multirow{ 2}{*}{70.5}&\multirow{ 2}{*}{70.5}&\multirow{ 2}{*}{71.3}\\
         Online\\
         \midrule
    \end{tabular}
    \caption{The effect of batch size for TTT. We provide the Mean Top-1 Accuracy (\%) over all the corruptions in the ModelNet-40C dataset for different batch sizes used for TTT.  The accuracy for Source-Only baseline is 57.6\%. }
    \label{tab:batch_size_abl}
    \vspace{-0.3cm}
\end{table}

\subsection{Strides for TTT}
\begin{figure}
\vspace{-0.3cm}
    \centering
    \includegraphics[scale=0.45, trim = 10 15 0 0, clip]{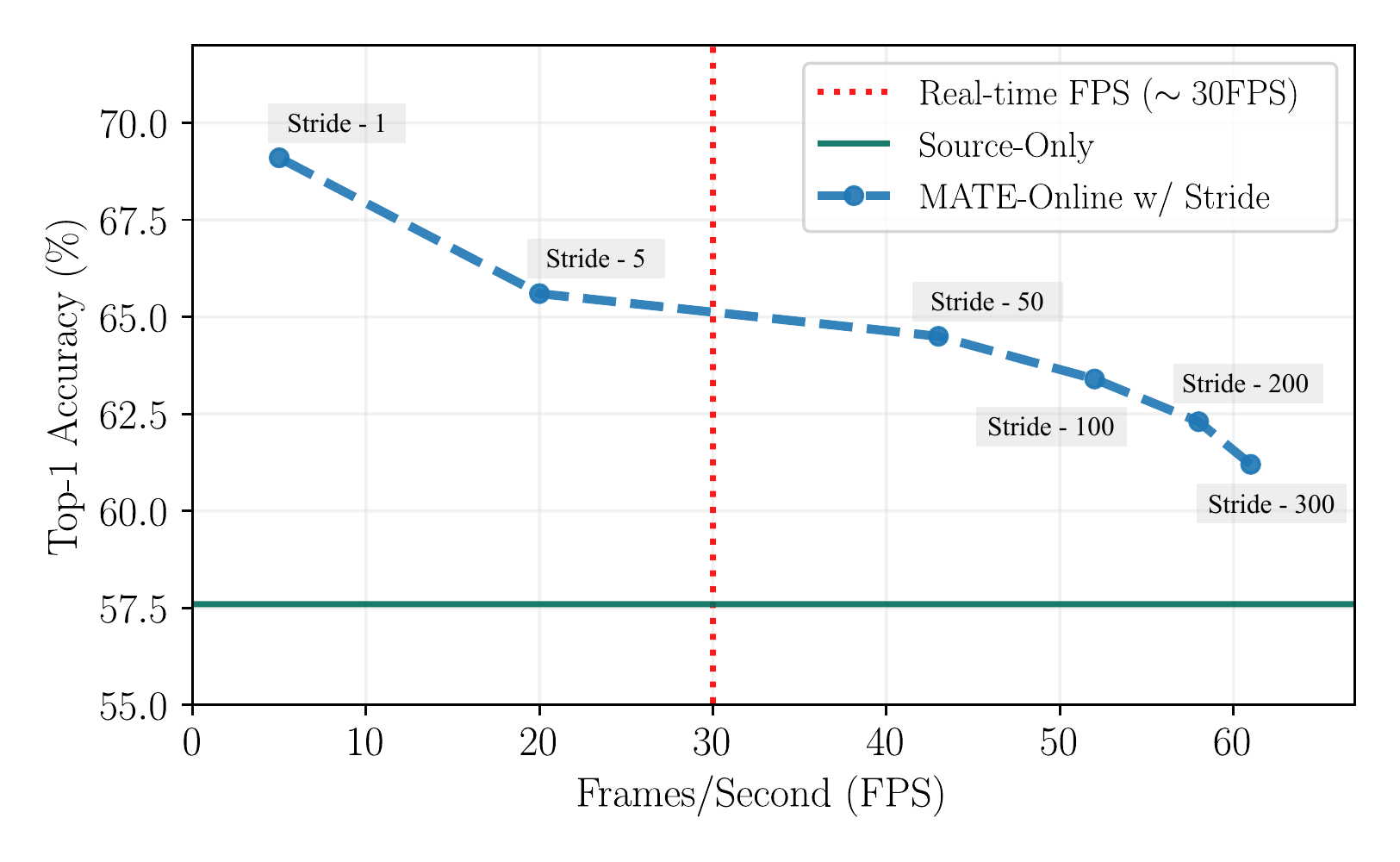}
    \caption{MATE can achieve real-time adaptation performance with only a minor performance penalty. Here, we report the Mean Top-1 Accuracy~(\%) over the 15 corruptions in the ShapeNet-C dataset for different adaptation strides. Strides represent the number of samples after which an adaptation step is performed.}
    \label{fig:ablation-stride}
    \vspace{-0.3cm}
\end{figure}

Some applications might require adaptation at test-time with minimum latency. 
For example, a test-time training method deployed in autonomous vehicles would ideally be required to adapt at a high frame-rate per second (FPS).
Thus, a test-time training method should be able to run with \emph{close to real-time} adaptation speed. 
Since most of the computation overhead for adaptation methods is during the backward pass, adapting to test samples sparsely should help to reduce the computation effort. 
In order to scratch the boundaries of our MATE for achieving a higher FPS, we design an experiment where we only adapt at test-time after a certain number of samples~(stride). 
Results for ShapeNet dataset in this scenario are provided in Figure~\ref{fig:ablation-stride}.
When performing an adaptation step after a stride, we find that our MATE can achieve close to real-time performance, with a minimum performance penalty.
For example, when we take a gradient step on every $5$-th sample, MATE can adapt at 20 FPS on an NVIDIA 3090 (for reference $30$ FPS is often considered as real-time~\cite{shi2022pillarnet}) with only~$\sim$$3$ percent-point drop in performance while comparing with the results obtained with a stride of $1$ (adapting on each incoming sample). 
We can even increase the stride up to $300$ and still achieve~$\sim$$3$ percent-point better performance than the Source-Only results, with an FPS of $62$. 
These results indicate the efficient nature of our MATE and its ability to show effective real-time adaptation performance. 
Results for ModelNet-40C in this adaptation protocol are provided in the supplementary.  
\vspace{-0.1cm}
\subsection{Batch Size for Test-Time Training}
MATE constructs a batch of $48$ from each point cloud encountered at test-time for adaptation. 
The point cloud in this batch is randomly masked and then masked patches are reconstructed.
Random masking helps us achieve a natural augmentation during test-time training.
To test the effect of our design choice on the test-time training performance, we experiment with different batch sizes on the ModelNet-40C dataset. 
These results are provided in Table~\ref{tab:batch_size_abl}. 
Surprisingly, for batch size of $1$, test-time adaptation performance falls below Source-Only but is $8.8$ percent-point better than Source-Only for the batch size of $2$. 
We also see that batch size larger than $8$ achieve minor gains, thus it could be a resource-efficient alternative. 

\begin{table}
\setlength\tabcolsep{3.0pt}
    \centering
    \small
    \begin{tabular}{l cccccc}
    & Source & JT & DUA & TTT-Rot & MATE-S & MATE-O \\
    \midrule
    Comb - 1& 33.9&36.7&{42.6}&34.3&\underline{47.7} &\bfseries{55.7}\\
      Comb - 2&29.6&34.7&{40.6}&32.9&\underline{45.2} &\bfseries{51.4}\\
       Comb - 3&28.3&33.3&{41.5}&30.7&\underline{44.5} &\bfseries{52.5}\\
       \midrule
       Mean &30.6&34.8&41.6&32.6&\underline{45.8}&\bfseries{53.2}\\
       \midrule
    \end{tabular}
    \caption{Top-1 Mean Accuracy~(\%) for three different datasets constructed by combining~$2$ randomly chosen corruptions for each subsequent sample in the test-set of ModelNet-40. JT: Joint Training, MATE-S: MATE-Standard, MATE-O: MATE-Online}
    \label{tab:corruption-combination}
    \vspace{-0.4cm}
\end{table}


\subsection{Combination of Distribution Shifts}
In realistic scenarios there could be situations where the test sample might be corrupted with a combination of corruptions. 
Thus, a test-time training method should be able to cope with such scenarios as well. 
To test our MATE in such a scenario, we design an experiment where we randomly combine 2 corruption types (from the ModelNet-40C benchmark) for each sample in the test set of ModelNet-40 and create $3$ such datasets. 
To generate these datasets, we ensure that all $15$ corruption types are selected for each dataset and for each sample $2$ corruptions are chosen randomly from the set of $15$ corruptions. 
We test our MATE and other baselines on these datasets and provide the results in Table~\ref{tab:corruption-combination}. 
We see that MATE can effectively adapt to this scenario as well and outperforms other baselines by a considerable margin. 
DUA fares better than TTT-Rot, because DUA does not use any geometric information, which is another indication that rotation prediction might not be a suitable test-time training objective for 3D point clouds. 

\subsection{Limitation} 
In this paper we propose the first TTT method for 3D point cloud data. 
To this end, we tested our MATE rigorously for the point cloud classification task.
Focusing on this task we were able to show that
masked autoencoders can provide extremely powerful self-supervisory signal for this task. 
However, application of TTT to other downstream tasks is out-of-scope for this work and thus we leave it for future exploration. 

\section{Conclusion}
\label{sec:conclusion}
Test-time training approaches designed for the 2D image domain can often degrade significantly if naively applied to the 3D data, requiring specialized 3D-specific designs.  
To this end, we are the first to propose a 3D test-time training method, MATE. 
We show that masked autoencoding is a powerful self-supervised auxiliary objective, which can make the network robust to various kinds of distribution shifts occurring in 3D point clouds.
Our MATE, is computationally cheap and can also run in real-time adaptation scenarios while achieving significant performance gains. 
\newpage

\appendix
\section*{Supplementary material}
In the following, we present the detailed Algorithm for MATE~(Section~\ref{sec:algorithm}), provide specifics for all the distribution shifts~(Section~\ref{sec:distribution-shifts-details}), present experiments on ModelNet-40C achieving real-time test-time training (TTT) (Section~\ref{sec:real-time-ttt}) and show correlation between TTT and the auxiliary task of MAE reconstruction (Section~\ref{sec:cls-and-recon}).

\section{Algorithm}
\label{sec:algorithm}
In~Algorithm. \ref{alg:pseudo}, we provide the detailed algorithm for our MATE, which consists of three phases: Joint training, Test-time training, and Online evaluation.

\section{Details about Distribution Shifts}
\label{sec:distribution-shifts-details}
We use the corruption benchmark~\cite{sun2022benchmarking} to introduce $15$ different types of commonly occurring
distribution shifts on the test sets of the point cloud datasets we use for evaluation in our main manuscript.
A description of these distribution shifts is provided as follows:
\begin{itemize}[nosep,label=-,wide]
\item  \emph{Uniform noise}: Random noise is added to each point in a point cloud, where the amount of noise is based on a uniform distribution and lie within a range of $\pm$0.05. 
\item \emph{Gaussian noise}: Points are randomly perturbed and the amount of noise is based on a Gaussian (normal) distribution with values in range of $\pm0.03$. 
\item  \emph{Background noise}: Randomly add ($\frac{Number\,of\,Points}{20}$) points with values in the range of $\pm1$ in the bounding box of the point cloud.

\item  \emph{Impulse noise}: 
Add a value in the range of $\pm0.1$ to a subset of the total number of points in the pointcloud.

\item  \emph{Upsampling}: Additional points are added by duplicating the existing points in a point cloud. 

\item  \emph{RBF}: The point clouds are deformed based on the Radial Basis Function~\cite{10.1080/10618562.2014.932352}. 
\item  \emph{Inverse\_RBF}: To generate this shift, the Radial Basis Function and the resulting splines are inverted. 
\item  \emph{Local\_Density\_Decrease}: To generate this distribution shift, $5$ local cluster centers and their $100$ closest neighbors are chosen. 
Further, their point density is decreased by deleting $\frac{3}{4}$ of the points inside the clusters.
\item  \emph{Local\_Density\_Increase}: 
Choose $5$ local cluster centers with $100$ closest neighbors. 
Then, keep these clusters, but randomly sample the rest of the pointcloud again with the original number of points. 
This results in double the density in the clusters, in comparison to the rest of the point cloud.
\begin{figure}
    \centering
    \includegraphics[scale=0.5]{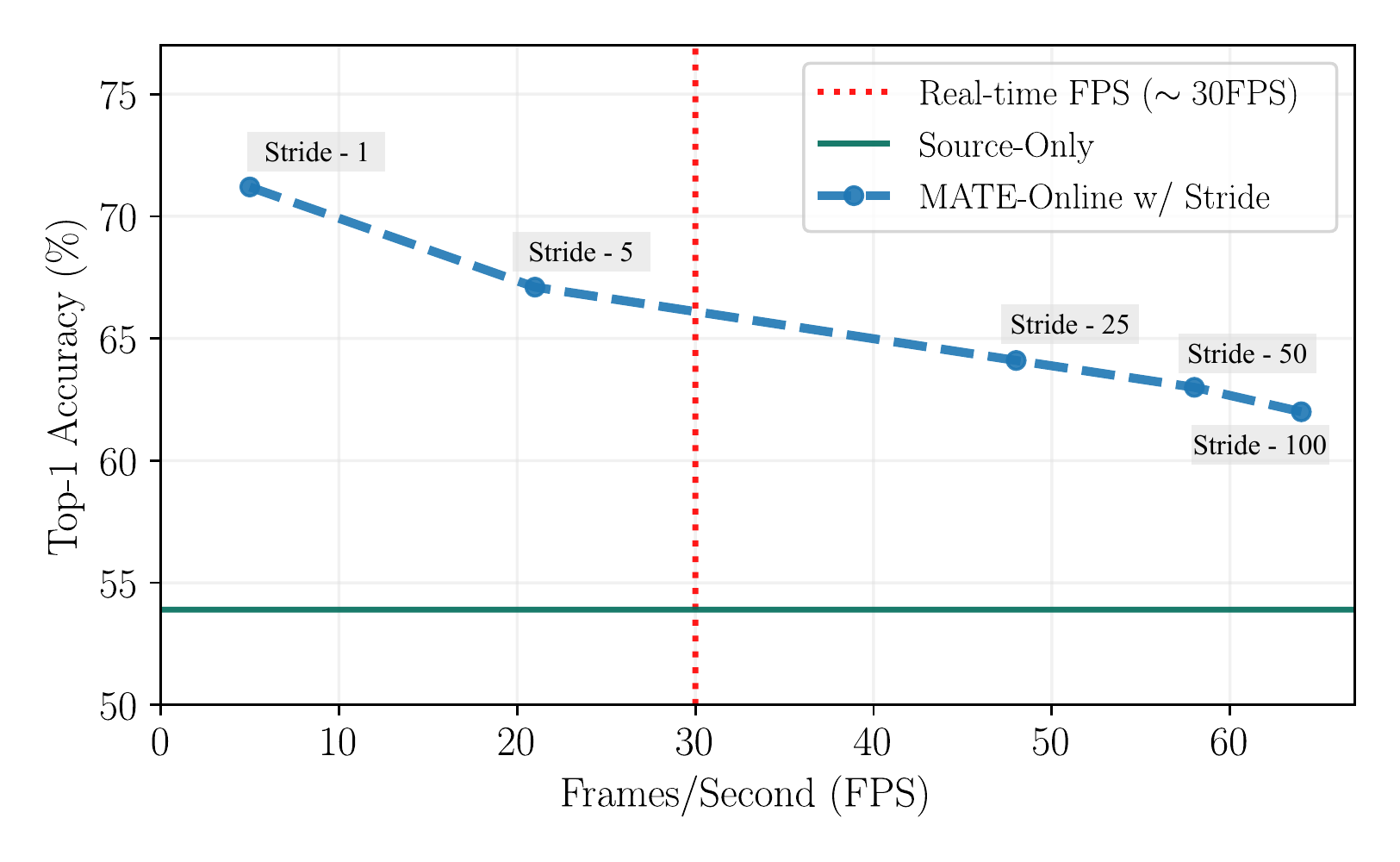}
    \caption{MATE can achieve real-time adaptation performance by only sacrificing some percent-points. Here, we report the Mean Top-1 Accuracy~(\%) over the 15 corruptions in the ModelNet-40C dataset for different adaptation strides. Strides represent the number of samples after which an adaptation step is performed.}
    \label{fig:stride-modelnet}
\end{figure}

\begin{figure*}
\begin{minipage}{0.48\textwidth}
    \centering
    \includegraphics[scale=0.5]{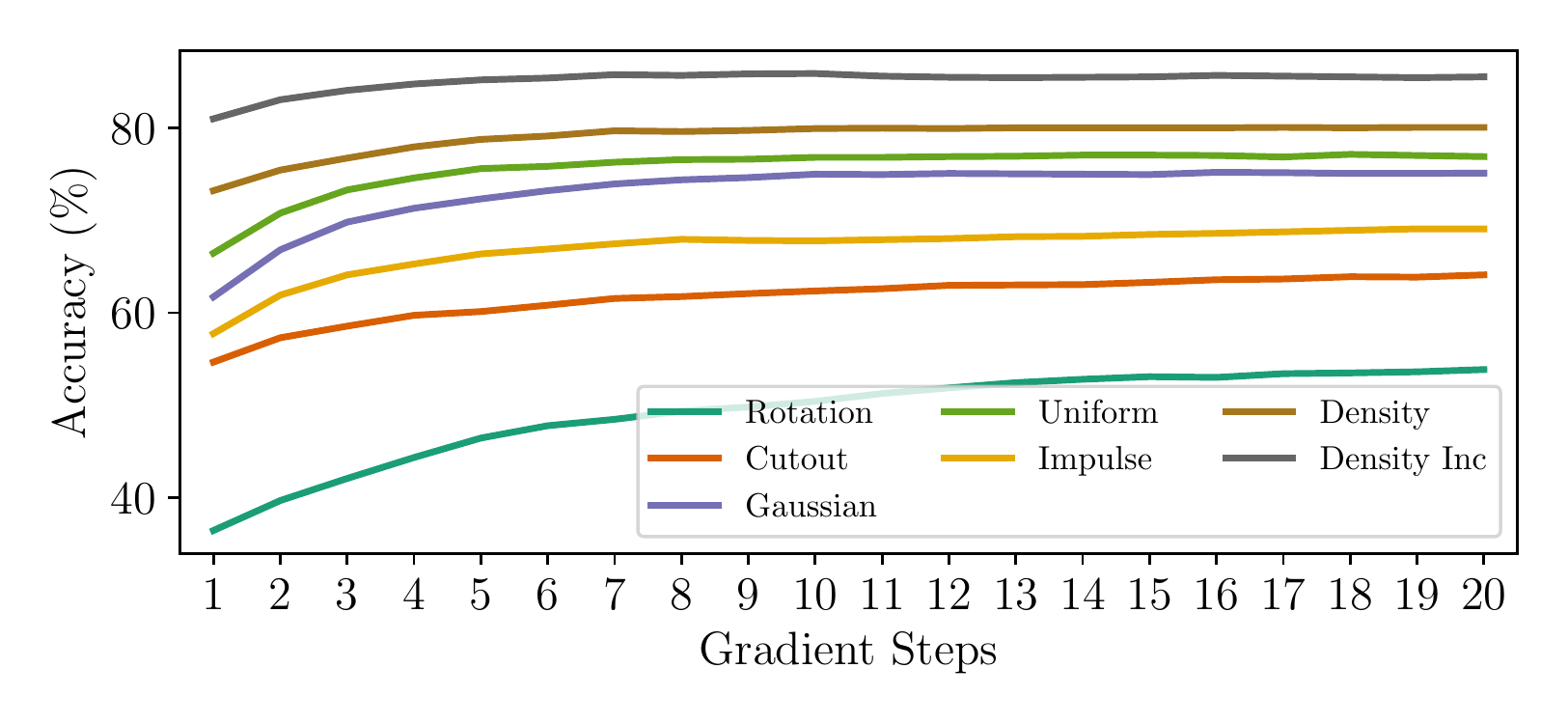}
    \includegraphics[scale=0.5]{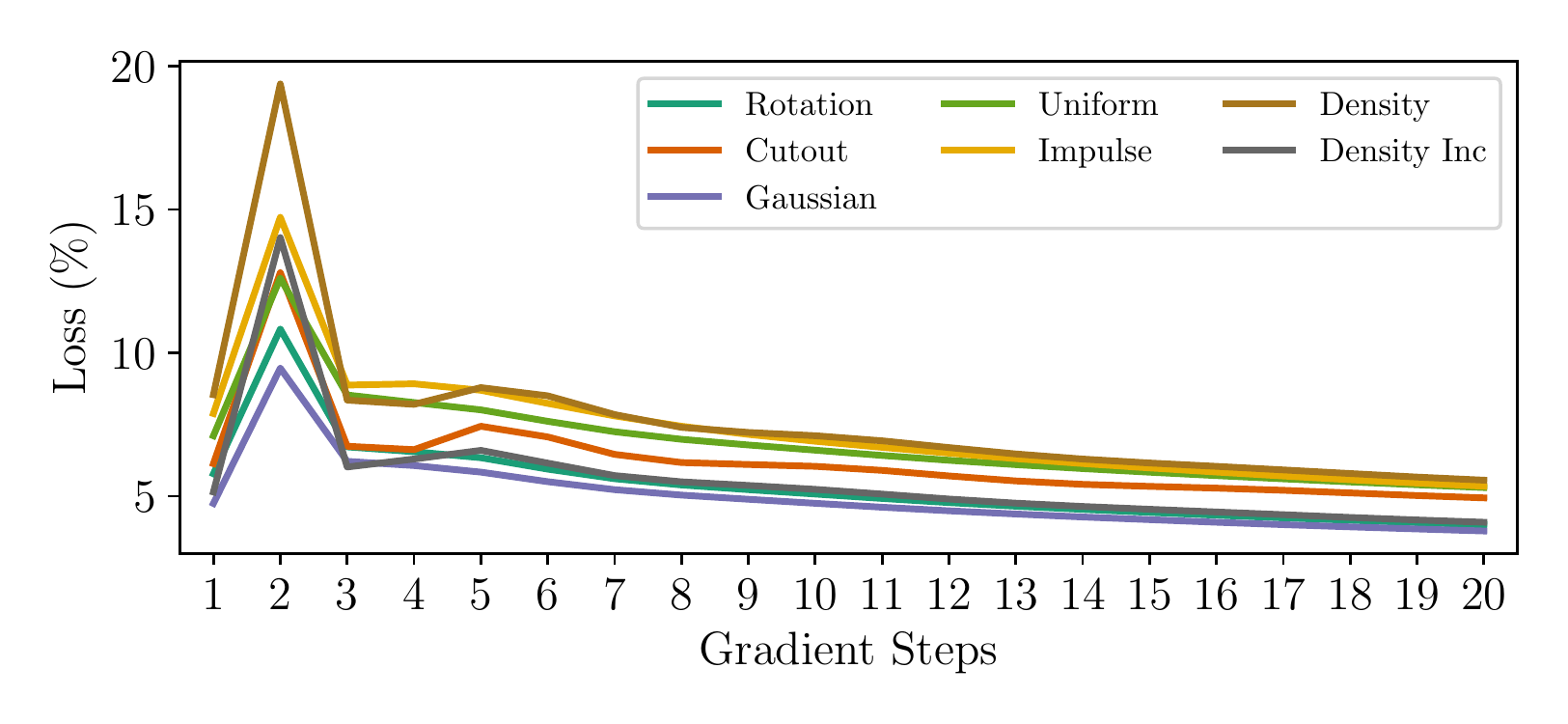}
\end{minipage}
\begin{minipage}{0.48\textwidth}
    \centering
    \includegraphics[scale=0.5]{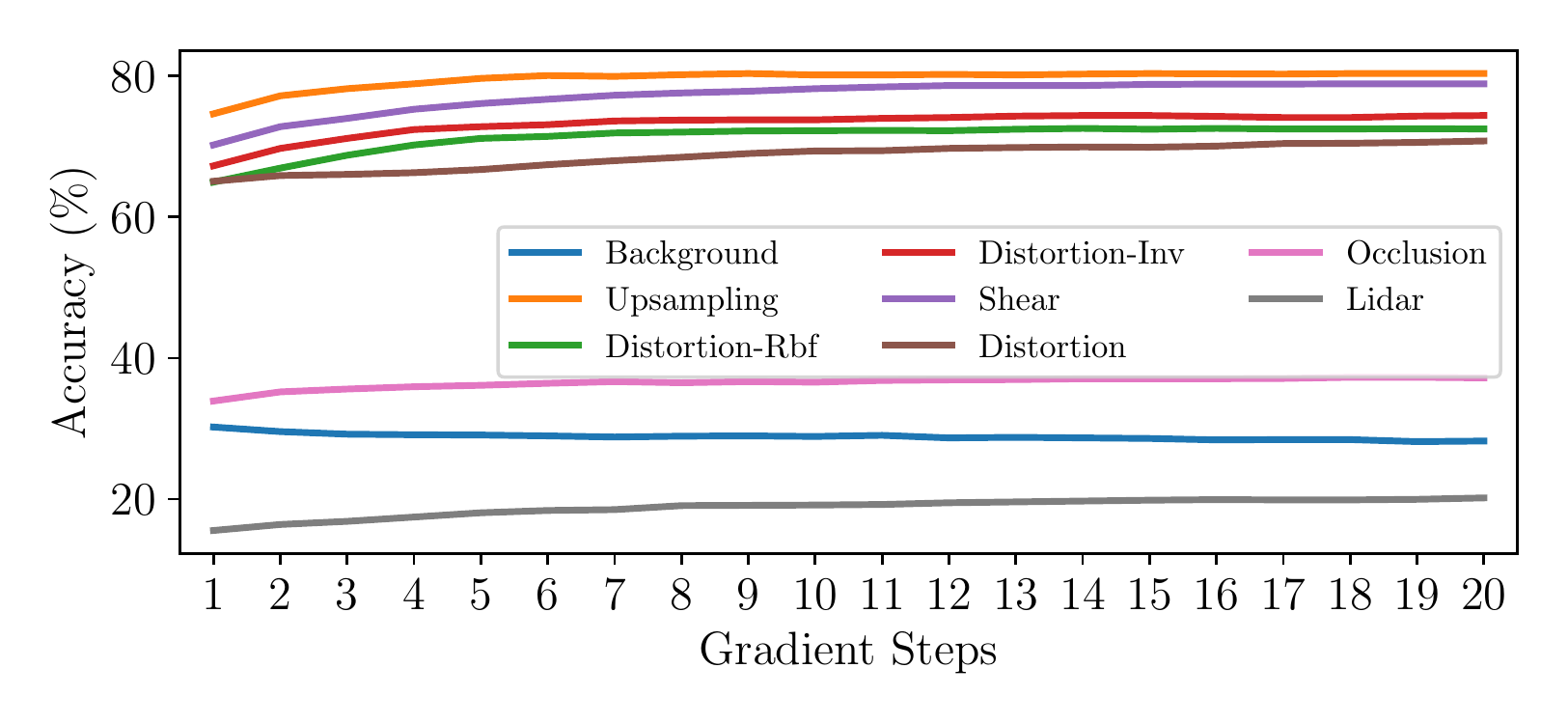}
    \includegraphics[scale=0.5]{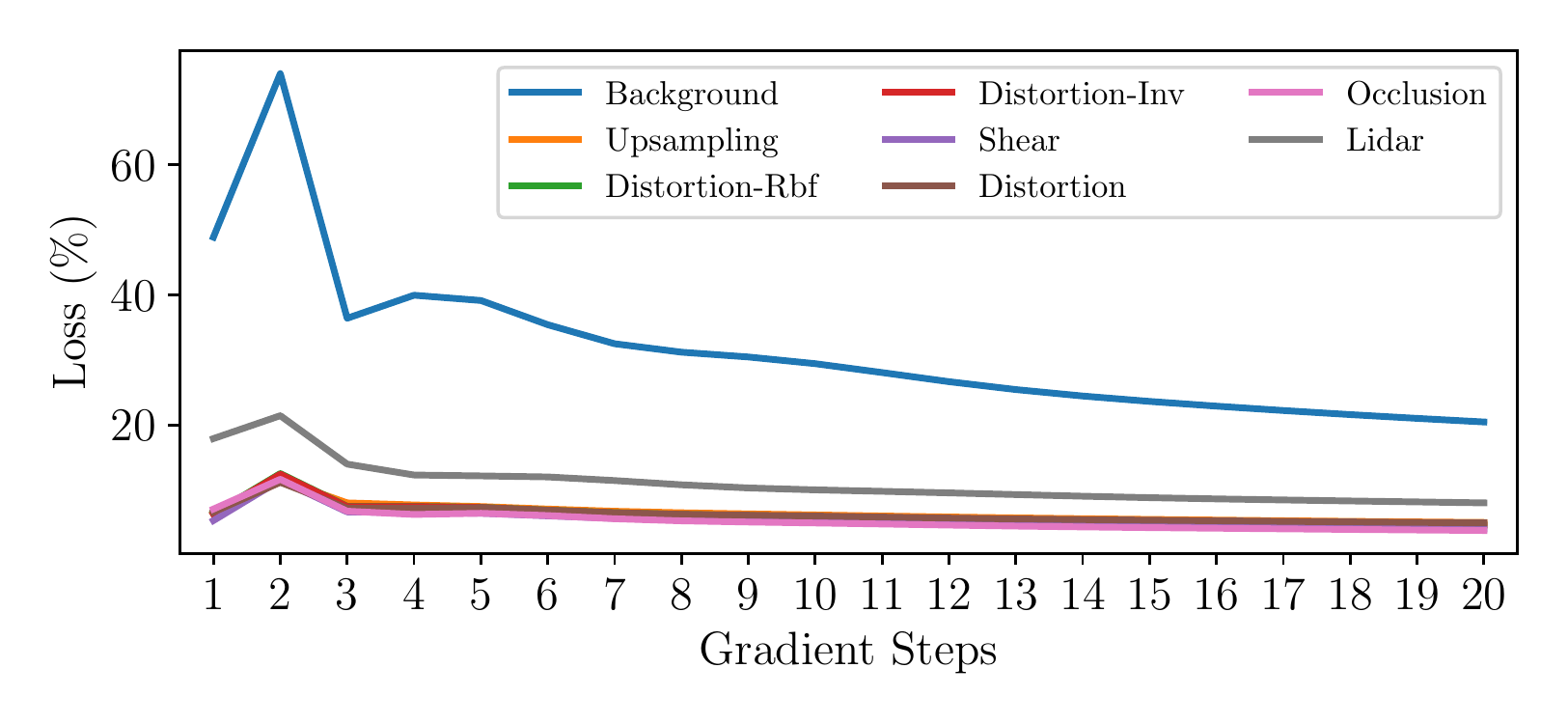}
\end{minipage}
\caption{Accuracy (Top) and Reconstruction Loss (Bottom) for all corruption in the ModelNet-40C at each adaptation step for MATE-Standard. To avoid clutter, we split the different corruptions into two plots (left and right).}
\label{fig:loss-acc-corelation}
\end{figure*}
\begin{figure*}
    \centering
    \includegraphics[scale=0.85]{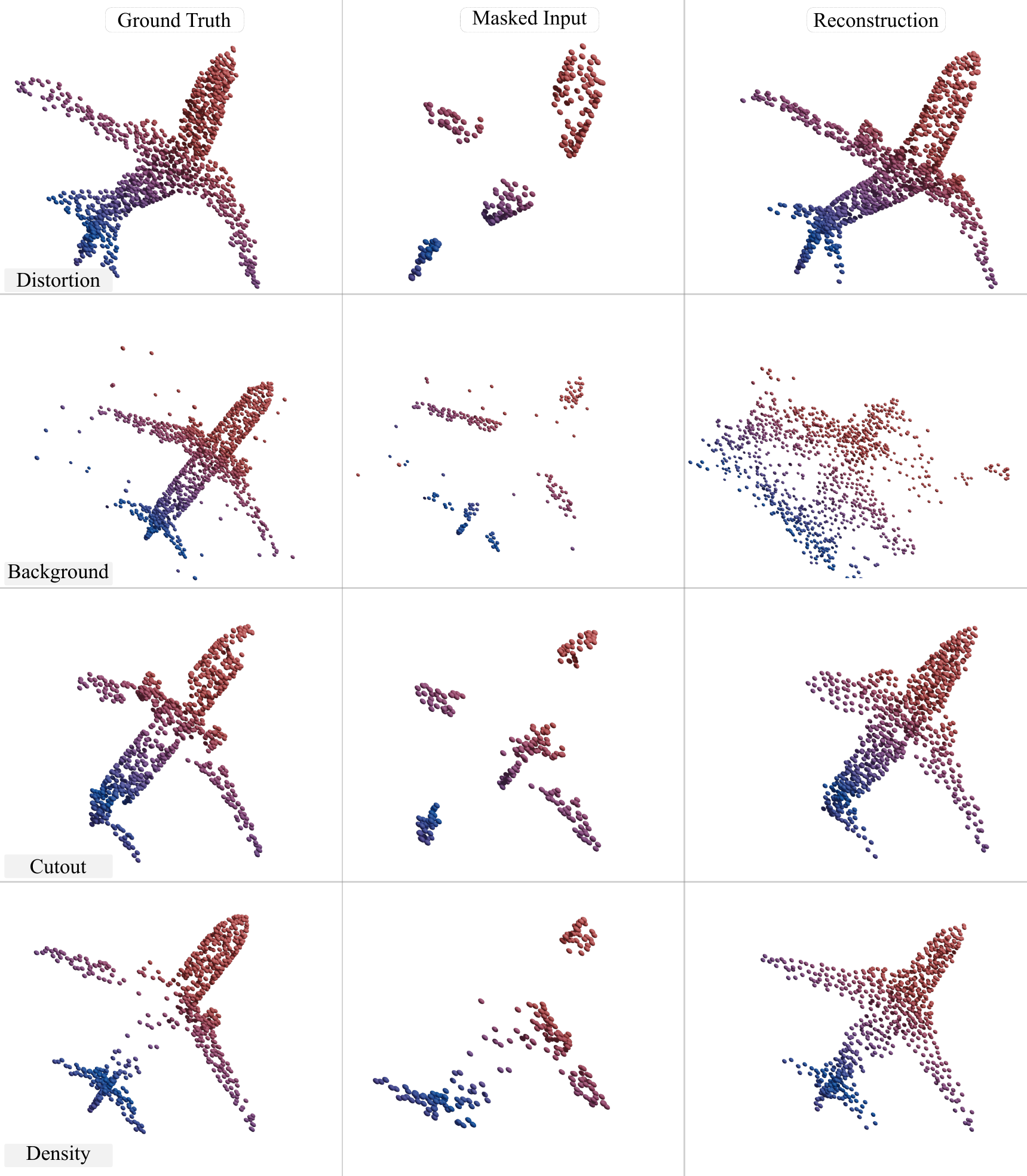}
    \vspace{0.2cm}
    \caption{Reconstruction results for MATE-Standard at the $20$-th gradient step for adaptation at test-time. We plot the out-of-distribution test sample for adaptation (left), 10\% input visible tokens (center) and the corresponding reconstruction output (right) for four corruptions in the ModelNet-40C dataset. }
    \label{fig:recon-org}
\end{figure*}
\item \emph{Shear}: Randomly compress and stretch the point cloud on the xy-plane. 
Here, the points get multiplied by values in range of $\pm$0.25 for each dimension.
\item \emph{Rotation}: Rotate all three spatial dimensions of a point cloud by a random angle in range of $\pm$15$^{\circ}$. 

\item  \emph{Cutout}: To simulate cutout, generate $5$ local clusters with $100$ closest neighbors and remove these clusters from the original point cloud.

\item  \emph{FFD}: For this distribution shift, the Free-Form Deformation (FFD)~\cite{10.1145/15922.15903} is used. 
The point cloud is enclosed in a box consisting of splines which are defined by control points. 
The control points are shifted to deform the point cloud.
A total of $125$ control points are used with a deformation distance in a range of $\pm0.5$.

\item  \emph{Occlusion}: Occluded points are deleted by using raytracing from a random camera position. 
For this operation, precomputed meshes~\cite{zhou2018open3d} are used. 
\item  \emph{LiDAR}: Point clouds are simulated as if they are generated from a LiDAR sensor.
In addition to occlusion, inaccuracies based on reflections and noise are added. 

\end{itemize}

\begin{algorithm}[hbt!]
\DontPrintSemicolon
\vspace{-0.5mm}
\caption{Algorithm for MATE}
\label{alg:pseudo}
\KwIn{(Training data $\mathcal{S}=\{(\mathcal{X}, \mathcal{Y})\}$, Single out-of-distribution point-cloud $\tilde{\mathcal{X}}$)}
\Begin
{
    Define the network with encoder $E$, decoder $D$, prediction head $P$, 
    classifier head $C$ \\
    Define the masking ratio~$m$, batch size~$b$,
    stride~$s$ 
    and gradient steps~$k$\\ 
    \textcolor{teal}{\# Joint Training.} \\
    \For{multiple epochs}
        {
            $\mathcal{X}^{v}$ = \textcolor{violet}{point-masking}$(\mathcal{X}, m)$ \\
            $L$ = $CE(C \circ E(\mathcal{X}^v), \mathcal{Y})$ + $CD(P \circ D \circ E(\mathcal{X}^{v}), \mathcal{X})$~(Eq. (\textcolor{red}{2, 3}))\\
            $L$.backward() \\
            optimizer.step()
        }
    \textcolor{teal}{\# Test-Time Training \& Online Evaluation.} \\
    \For{idx, $\tilde{\mathcal{X}}$ in loader}
        {
            \If{idx \% $s$ == 0}{
                $L_{TTT}$ = 0 \\
                \For{$k$ iterations}
                    {
                        $\tilde{\mathcal{X}^{v}}$ = [\textcolor{violet}{point-masking}$(\tilde{\mathcal{X}}, m)$ for \_ in \textcolor{blue}{range}($b$)] \\
                        $L_{TTT}$ += $CD(P \circ D \circ E(\tilde{\mathcal{X}}^{v}), \tilde{\mathcal{X})}$ (Eq. (\textcolor{red}{4})) \\
                    }
                $L$ = $L_{TTT}$.mean() \\
                $L$.backward() \\
                optimizer.step()
            }
            Evaluate $C \circ E(\tilde{\mathcal{X}}^v)$
        }
}
\end{algorithm}

\section{Real-time Test-time Training}
\label{sec:real-time-ttt}
In the main manuscript (Figure 3), we provide results for MATE-Online while adapting sparingly to the test samples in ShapeNet-C dataset. 
We see that, while adapting sparingly on the test data,~\ie only back-propagating gradients after a certain number of samples (stride), our MATE can still achieve strong performance gains and can even match the real-time FPS ($30$), with only a minimum penalty on accuracy. 
Here, in Figure~\ref{fig:stride-modelnet}, we provide the results with different strides for the ModelNet-40C dataset, which is $\sim4\times$ 
smaller than the ShapeNet-C dataset. 
While adapting sparingly, we see that, similar to the results on the large-scale ShapeNet-C, our MATE can also achieve close to \textit{real-time} performance by dropping only a few percent-points as compared to adapting on each sample. 
For example, with a stride of $5$ (adapting on every $5$-th sample), our MATE drops only $\sim3$ percent-points as compared to the results with stride-1 (adapting on each incoming sample), while obtaining an FPS of $21$.

\section{Classification and Reconstruction}
\label{sec:cls-and-recon}
At test-time, MATE adapts to each out-of-distribution (OOD) test sample by using the self-supervised reconstruction task as an auxiliary objective, leveraging masked autoencoders~\cite{pang2022masked}.
As each OOD sample is encountered, the network is adapted by the auxiliary self-supervised loss. 
This loss is an $l_2$ Chamfer distance between the reconstructed masked tokens and the corresponding ground truth tokens from the original OOD test sample. 
After adapting the network by back-propagating the gradients obtained from the auxiliary loss, the OOD sample is evaluated. 
In the main manuscript, we see that our test-time training methodology achieves strong performance gains on a variety of datasets for object classification in 3D point clouds.
Naturally, the question arises -- `How a self-supervised task,~\ie~reconstruction task, can help to adapt the network for a seemingly unrelated task, like object classification?' 
Through our experiments, we find that there is a correlation between the reconstruction task and the classification task and that is the reason for the improvement in classification accuracy by simply reconstructing the corrupted (OOD) test sample at test-time. 
We find this correlation empirically through two procedures, detailed in the following. 

\subsection{Loss and Accuracy}
In the main manuscript, we test our MATE in two test-time
training variants, described in Section 3.5, here we again provide a brief description in the interest of keeping the reading flow:

\paragraph{MATE-Standard} assumes access to a single sample at test-time for adaptation. 
In order to adapt the network on this single sample, we take multiple gradient steps~(\ie~20) for test-time training.
After adaptation on each sample, the network weights are re-initialized for adaptation on the next sample. 

\paragraph{MATE-Online} assumes access to a stream of data for adaptation and the network updates are accumulated after adaptation on each sample in the stream.
For this adaptation variant, we only take a single gradient step on each OOD test sample. 

In Figure~\ref{fig:loss-acc-corelation}, we plot the Top-1 Accuracy on ModelNet-40C and the corresponding reconstruction loss at each gradient step for MATE-Standard.
Please note that these results are plotted by taking the average of the accuracy over all the samples in the test set of ModelNet-40C at each gradient step.
From the results it is evident that as the reconstruction loss decreases after each gradient step, the corresponding accuracy increases. 
This shows that as the model becomes better at reconstructing the OOD test sample, the classification performance
is influenced in a positive way.  
We also see that 
there is a spike in the reconstruction loss during the initial update step. 
We hypothesize that this is because of the sudden distribution shift which is encountered at test-time, since the model is initially trained on clean point clouds. 
However, with more adaptation steps for test-time training, it slowly gets better at reconstructing the OOD sample.  

Furthermore, an interesting result is that of the \emph{Background} corruption. 
In the main manuscript, while listing the results for ModelNet-C (Section 4.4) we found that for the background corruption TTT-Rot~\cite{sun2020ttt} fares better than our MATE. 
From Figure~\ref{fig:loss-acc-corelation}, we see that for Background corruption the reconstruction loss is highest among all the other corruptions, that can be one of the reasons why MATE cannot perform well on this corruption. 
This also gives us an indication of the correlation between the reconstruction and the classification loss. 
To further investigate the background corruption and find more answers behind correlation of the two tasks, we visualize the reconstruction results next. 
\subsection{Reconstruction Results}
We further analyzed the reconstruction results for different corruption types to get a deeper insight in to the correlation of the MAE reconstruction and the classification task. 
We find that for TTT with MATE-Standard, after $20$ gradient steps, the reconstruction for the Background corruption is the worst as compared to reconstruction of other corruption types. 
We visualize these results for a few corruptions in the ModelNet-40C dataset for the $Airplane$ class in Figure~\ref{fig:recon-org}.
Reconstructions from the remaining corruptions also follow a similar pattern. 
Since MATE does not perform optimally for the Background corruption, this gives us an indication of the correlation between the auxiliary self-supervised reconstruction task and the downstream classification task. 
To conclude, our results show that if the auxiliary self-supervised reconstruction task is able to reconstruct the input corruption type optimally, MATE shows strong performance gains, which is an indication that these two tasks are correlated with each other.  
{\small
\bibliographystyle{ieee_fullname}
\bibliography{main}
}

\end{document}